\documentclass{article} 
\usepackage{iclr2025_conference,times}

\usepackage{amsmath,amsfonts,bm}

\def\eqref#1{equation~\ref{#1}}

\def\1{\bm{1}}

\DeclareMathAlphabet{\mathsfit}{\encodingdefault}{\sfdefault}{m}{sl}
\SetMathAlphabet{\mathsfit}{bold}{\encodingdefault}{\sfdefault}{bx}{n}

\usepackage{amssymb,mathtools,amsthm}
\usepackage{algorithm,algorithmic}
\usepackage{bbold}

\usepackage[colorlinks,citecolor=mydarkblue]{hyperref}
\usepackage{url}
\usepackage{enumitem}
\usepackage{xspace}
\usepackage{multirow}
\usepackage{booktabs}
\usepackage{graphicx}
\usepackage{caption}
\usepackage{subcaption}
\usepackage{wrapfig}
\usepackage{pifont}

\definecolor{mydarkblue}{rgb}{0, 0, 0.5}
\definecolor{kmycolor}{rgb}{0.8,0.2,0.4}
\definecolor{forestgreen}{rgb}{0.0, 0.5, 0.0}

\newcommand{\method}{COrAL\xspace}

\newcommand{\deltavaldown}[2]{$#1$\color{red} $\downarrow$\textsubscript{$#2$}}
\newcommand{\deltavalup}[2]{$#1$\color{forestgreen} $\uparrow$\textsubscript{$#2$}}

\title{\method: Order-Agnostic Language Modeling for Efficient Iterative Refinement}

\author{%
  Yuxi Xie$^{1,2}$\thanks{Work done during visiting at UCSB.} \quad Anirudh Goyal$^{3}$ \quad Xiaobao Wu$^{2,4}$\footnotemark[1] \quad Xunjian Yin$^{2}$\footnotemark[1] \quad Xiao Xu$^{1}$ \\
  \textbf{Min-Yen Kan}$^{1}$ \quad  \textbf{Liangming Pan}$^{5}$ \quad \textbf{William Yang Wang}$^{2}$ \\
  $^1$ National University of Singapore \quad
  $^2$ University of California, Santa Barbara \quad
  $^3$  Mila\\
  $^4$ Nanyang Technological University \quad
  $^5$ University of Arizona \\
  \texttt{xieyuxi@u.nus.edu} \quad \texttt{william@cs.ucsb.edu}
}

\newcommand{\yuxi}[1]{\textcolor{teal}{$_{yuxi}$[#1]}}

\def\gcheck{\textcolor{forestgreen}{\ding{51}}}
\def\rcross{\textcolor{red}{\ding{55}}}

\iclrfinalcopy 
\begin{document}

\maketitle

\begin{abstract}
Iterative refinement has emerged as an effective paradigm for enhancing the capabilities of large language models (LLMs) on complex tasks. However, existing approaches typically implement iterative refinement at the application or prompting level, relying on autoregressive (AR) modeling. The sequential token generation in AR models can lead to high inference latency. 
To overcome these challenges, we propose \underline{C}ontext-Wise \underline{Or}der-\underline{A}gnostic \underline{L}anguage Modeling (\method), which incorporates iterative refinement directly into the LLM architecture while maintaining computational efficiency. Our approach models multiple token dependencies within manageable context windows, enabling the model to perform iterative refinement internally during the generation process. Leveraging the order-agnostic nature of \method, we introduce sliding blockwise order-agnostic decoding, which performs multi-token forward prediction and backward reconstruction within context windows. This allows the model to iteratively refine its outputs in parallel in the sliding block, effectively capturing diverse dependencies without the high inference cost of sequential generation.
Empirical evaluations on reasoning tasks demonstrate that \method improves performance and inference speed, respectively, achieving absolute accuracy gains of $4.6\%$ on GSM8K and $4.0\%$ on LogiQA, along with inference speedups of up to $3.9\times$ over next-token baselines. Preliminary results on code generation indicate a drop in 
pass rates due to inconsistencies in order-agnostic outputs, highlighting the inherent quality--speed trade-off.
Our code is publicly available at \href{https://github.com/YuxiXie/COrAL}{https://github.com/YuxiXie/COrAL}.\looseness=-1

\end{abstract}

\section{Introduction}\label{sec:intro}
\begin{wrapfigure}{r}{.47\textwidth}
    \centering
    \vskip -.1in
    \includegraphics[width=.47\textwidth]{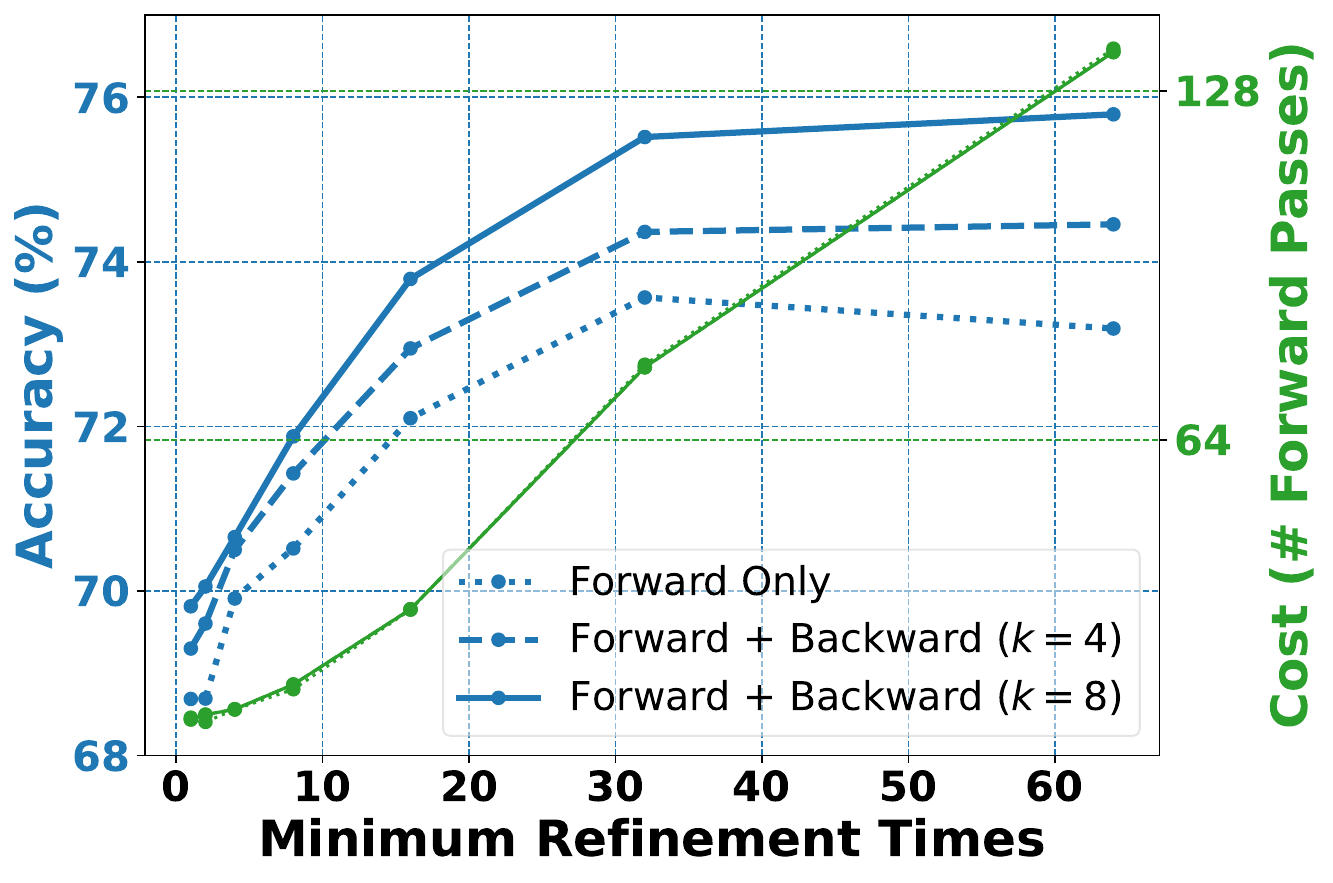}
    \vskip -.1in
    \caption{Scaling of performance and inference cost on GSM8K with increasing the minimum refinement times for each output position. $k$ represents the backward context window size. We set the decoding block size as $b=64$.}
    \label{fig:it}
    \vskip -.2in
\end{wrapfigure}
Large Language Models (LLMs) have recently achieved remarkable success across a wide range of tasks~\citep{DBLP:conf/nips/BrownMRSKDNSSAA20, DBLP:journals/corr/abs-2307-09288, DBLP:journals/corr/abs-2303-08774, DBLP:journals/corr/abs-2407-21783}, such as mathematical problem-solving, logical reasoning, and programming~\citep{DBLP:conf/iclr/YuJSYLZKLWL24,DBLP:conf/emnlp/PanAWW23,DBLP:conf/nips/SchickDDRLHZCS23,DBLP:journals/corr/abs-2308-12950}.
Strategies that enable LLMs to learn from previous mistakes and iteratively refine their outputs have been particularly effective, achieving human-level performance and transforming both academic research and industrial applications~\citep{DBLP:journals/tacl/PanSXNWW24,ye2024physicslanguagemodels22,openai-o1}.
These iterative refinement approaches incorporate feedback---either external or internal---as supervision signals during training~\citep{DBLP:conf/nips/ZelikmanWMG22,DBLP:conf/emnlp/0001GHW00023,DBLP:conf/nips/ShinnCGNY23,DBLP:conf/iclr/LightmanKBEBLLS24,DBLP:journals/corr/abs-2405-00451}, or by developing prompting frameworks that guide the model toward improved generations through methods like guided search or self-refine~\citep{DBLP:conf/nips/YaoYZS00N23,DBLP:conf/nips/XieKZZKHX23,DBLP:conf/nips/MadaanTGHGW0DPY23}.\looseness=-1

Despite their effectiveness, these approaches predominantly rely on autoregressive (AR) LLMs, which generate text by predicting the next token in a fixed left-to-right order using causally masked Transformers~\citep{radford2018improving}. This sequential generation process inherently limits the model's ability to capture dependencies spanning beyond the immediate next token, especially those that require backward context~\citep{DBLP:conf/iclr/HuJEKLBM24}. Moreover, the sequential nature of AR models leads to high inference latency, resulting in computational inefficiency for long sequences~\citep{DBLP:conf/icml/CaiLGPLCD24}.

\begin{figure}[t]
    \centering
    \includegraphics[width=\textwidth]{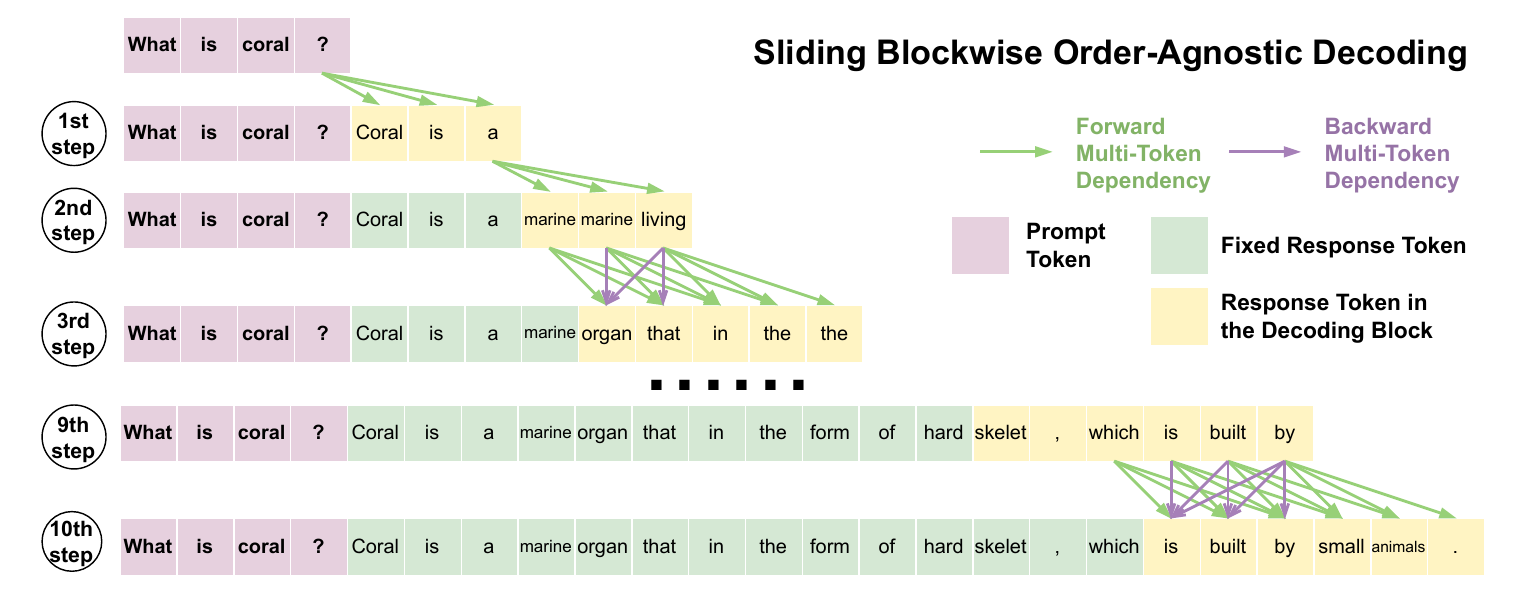}
    \vskip -.15in
    \caption{\textbf{Sliding Blockwise Order-Agnostic Decoding}. \method performs multi-token prediction and refinement in the sliding block with context window size $k=\!3$ and block size $b\!=\!6$.}
    \label{fig:decoding}
    \vskip -.1in
\end{figure}


To address these limitations, researchers have explored order-agnostic architectures that enhance representation learning and accelerate inference. Previous studies mainly focus on two solutions: permutation-based AR and non-autoregressive (NAR) modeling, but each has its own strengths and limitations.
For instance, permutation-based models propose diversity-enhanced pretraining objectives that predict multiple subsequent tokens in various orders to capture richer dependencies~\citep{DBLP:conf/nips/YangDYCSL19,DBLP:conf/icml/ZhangDHWW24}.
Similarly, NAR models generate tokens in parallel, significantly reducing inference time~\citep{DBLP:conf/iclr/Gu0XLS18}.
However, conventional NAR models often struggle with tasks involving variable-length generation and complex token dependencies, leading to degraded text quality. As a result, these models are typically task-specific and require additional mechanisms to ensure consistency~\citep{DBLP:conf/nips/GuiSMZC023,DBLP:journals/corr/abs-2403-02249}.
Inspired by the success of diffusion models in image generation~\citep{DBLP:conf/nips/AustinJHTB21}, recent efforts have adapted denoising techniques to generative language modeling as an iterative extension of NAR models~\citep{DBLP:conf/iclr/SavinovCBEO22,DBLP:conf/iclr/GongLF0K23}. While these methods improve efficiency, they still lag behind AR models regarding generation quality and generalizability.
Given the trade-offs among different models\footnote{We make conceptual comparison among different model architectures in Appendix~\ref{app:comp}.}, a pivotal question arises:\looseness=-1

\textit{Can we unify the strengths of denoising techniques with order-agnostic modeling to enhance the capabilities of AR-LLMs while mitigating their respective limitations}?

In this work, we propose \textbf{\underline{C}ontext-Wise \underline{Or}der-\underline{A}gnostic \underline{L}anguage Modeling} (\method), which combines the advantages of AR and order-agnostic modeling.
\method models token dependencies within manageable context windows, effectively balancing the capture of both local and long-range dependencies with computational efficiency. Through context-wise modeling, \method overcomes the limitations of fixed-order generation in AR models and the dependency modeling challenges in NAR models. 
Within each context window, \method models diverse dependencies in an order-agnostic manner, enhancing the model's ability to capture complex token relationships while maintaining computational efficiency.
Leveraging \method, we introduce \textbf{Sliding Blockwise Order-Agnostic Decoding}, which performs forward multi-token prediction and backward reconstruction simultaneously. As shown in Figure~\ref{fig:it}, this strategy enables the model to perform iterative refinement internally to scale up inference performance.
Additionally, to ensure that the model remains aware of target token positions without necessitating architectural changes, we apply a generalized Rotary Position Embedding (RoPE)~\citep{DBLP:journals/ijon/SuALPBL24} to the last layer of the Transformer. This positional encoding technique preserves target-aware representations, which are essential for effective order-agnostic generation and iterative refinement.

We conduct extensive experiments on reasoning tasks, including arithmetic computation and logical reasoning, to evaluate the effectiveness and efficiency of \method. Our empirical results show that \method not only improves performance but also significantly accelerates inference. Specifically, \method achieves absolute accuracy gains of $4.6\%$ on GSM8K and $4.0\%$ on LogiQA, along with inference speedups of up to $3.9$ times over next-token baselines.
These findings demonstrate that \method effectively captures dependencies within context windows while maintaining computational efficiency.
However, preliminary experiments on code generation reveal a decrease in pass rates due to inconsistencies in order-agnostic outputs, highlighting the inherent quality--speed trade-offs. This suggests that further refinements are necessary for tasks that require strict syntactic coherence.

\section{Context-Wise Order-Agnostic Language Modeling}\label{sec:method-train}
We present Context-Wise Order-Agnostic Language Modeling, a generalized AR framework that captures conditional textual distributions based on various orders in context windows.
We first review order-agnostic autoregressive language modeling and non-autoregressive modeling with iterative refinement as background, before describing \method{}'s objective and architecture.  
\begin{figure}[t]
\begin{center}
\includegraphics[width=\textwidth]{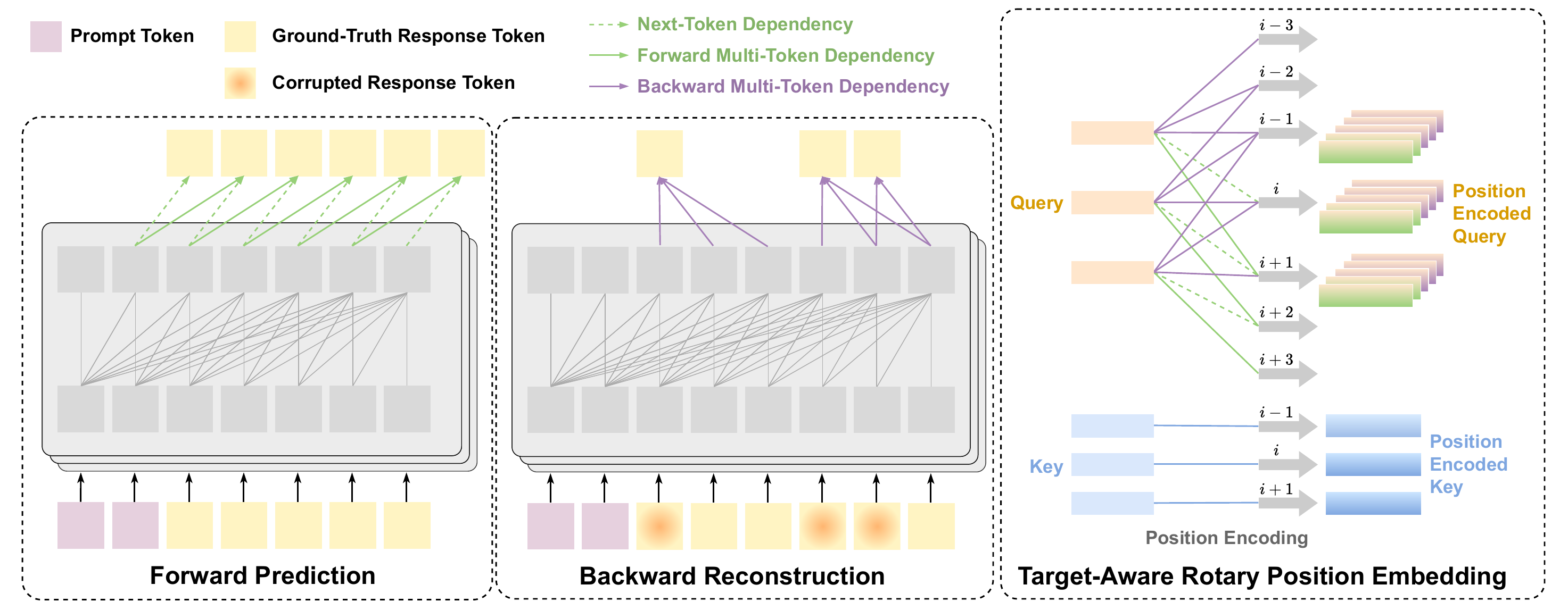}
\end{center}
\vskip -.15in
\caption{\textbf{Context-Wise Order-Agnostic Language Modeling}. We visualize the order-agnostic dependencies within a context window size $k=2$. For target-aware position encoding, we show how \method obtains query representations for multiple positions within a context window size $k=2$.}
\label{fig:model}
\vskip -.1in
\end{figure}

\subsection{Background}\label{method:pre}
Given a prompt $\bm{x}$ and a target sequence of $T$ tokens $\bm{y} = \{y_1, y_2, \cdots, y_T\}$, conventional AR models factorize the multivariate distribution $p(\bm{y}\mid\bm{x})$ into a product of univariate distributions using the probability chain rule:
\begin{equation}
    \vspace{-.065in}
    \log{p(\bm{y}\mid\bm{x})} = \sum_{t=1}^T\log{p(y_t\mid \bm{y}_{<t}, \bm{x})}
    \label{eq:ar}
\end{equation}
which requires $T$ iterative sampling steps to generate the sequence. In contrast, order-agnostic AR modeling generalizes this by modeling multiple possible orderings $\sigma\in \mathcal{S}_T$ of the sequence:
\begin{equation}
    \vspace{-.065in}
    \begin{aligned}
        \log{p(\bm{y}\mid\bm{x})} & = \log{\mathbb{E}_{\sigma\sim \mathcal{S}_T}p(\bm{y}\mid\bm{x},\sigma)} \\
        & \geq \mathbb{E}_{\sigma\sim \mathcal{S}_T}\log{p(\bm{y}\mid\bm{x},\sigma)} = \mathbb{E}_{\sigma\sim \mathcal{S}_T}\sum_{t=1}^T\log{p\left(y_{\sigma(t)}\mid\bm{y}_{\sigma(<t)}, \bm{x}\right)}
    \end{aligned}
    \label{eq:oa-ar}
\end{equation}
where $\mathcal{S}_T$ denotes the set of all possible permutations of the indices $\{1, 2, \cdots, T\}$. However, this permutation-based objective poses a significant optimization challenge and can lead to underfitting, as observed in prior works~\citep{DBLP:conf/nips/YangDYCSL19, DBLP:conf/iclr/HoogeboomGBPBS22}. 

On the other hand, NAR modeling~\citep{DBLP:conf/emnlp/LeeMC18} breaks the sequential dependency to accelerate inference. This approach applies sequence-level denoising steps, enabling parallel reconstruction of multiple tokens with iterative refinement to enhance generation quality. To equip the model with denoising capabilities, it employs $L$ intermediate latent variables $\{\bm{y}^{(1)}, \bm{y}^{(2)}, \cdots, \bm{y}^{(L)}\}$ and approximates their marginalization as follows:
\begin{equation}
    \vspace{-.065in}
    \log{p(\bm{y}\mid\bm{x})}
    \geq \sum_{t=1}^T\log{p(y_t\mid \bm{y}^{(L)}, \bm{x})} + \sum_{l=1}^L\sum_{t=1}^T\log{p(y_t^{(l)}\mid \bm{y}^{(l-1)}, \bm{x})} + \sum_{t=1}^T\log{p(y_t^{(0)}\mid \bm{x})}
    \label{eq:nar}
\end{equation}
where the latent variables are constrained to match the type of the target output $\bm{y}$. While previous studies demonstrate the efficiency of NAR modeling in specific tasks such as machine translation~\citep{DBLP:conf/iclr/Gu0XLS18, DBLP:conf/emnlp/GhazvininejadLL19, DBLP:conf/icml/KasaiCGG20}, its potential in language modeling remains underexplored. Moreover, the use of corrupted data for denoising and the assumption of token-wise independence in each reconstruction step in NAR models can introduce instability, often resulting in reduced text quality compared to their AR counterparts~\citep{DBLP:conf/iclr/SavinovCBEO22}.

\subsection{Objective: Context-Wise Order-Agnostic Autoregressive Modeling}\label{method:obj}
To address the above limitations in AR language modeling, we propose Context-Wise Order-Agnostic Language Modeling (\method), unifying token-level dependency modeling and sequence-level denoising to advance the capabilities of current LLMs. Previous order-agnostic modeling works attempt to capture various factorization orders involving long dependencies that are difficult to model and fit. In contrast, \method learns the orderless relationships within predetermined context windows. Built on the AR foundation, our \method framework leverages the superior capability of sequential language modeling in LLMs.

\method tackles the problem of generative language modeling by combining forward multi-token prediction with backward denoising in a context-wise and order-agnostic framework. Denoting the context window size as $k$\footnote{Without loss of generality, we can set different context window sizes for forward prediction and backward reconstruction in practice. Here, we present the objective with the same hyperparameter $k$ to avoid clutter.}, we model the conditional probability distribution of each target token by considering an ensemble of dependencies over all possible positions in the context:
\begin{equation}
    \vspace{-.065in}
    \log{p_{\theta}(\bm{y}\mid\bm{x})}
    \geq \sum_{t=1}^T\mathbb{E}_{i\in[t-k, t+k]}\mathbb{E}_{l\geq 0}\log{p_{\theta}(y_t\mid \bm{y}_{\leq i}^{(l)}, \bm{x})}
    \label{eq:oa-llm}
\end{equation}
where $\bm{y}^{(l)}$ represents an intermediate state of the target output sequence $\bm{y}$ during iterative refinement. The conventional AR modeling, in comparison, becomes a specific case where only the forward prediction with $k\!\!=\!\!1$, conditioned on previous tokens in the target sequence $\bm{y}$, is modeled. 

\paragraph{Forward Prediction and Backward Reconstruction.}
As shown in Figure~\ref{fig:model}, we decompose the order-agnostic objective into \textbf{forward prediction} and \textbf{backward reconstruction}. In forward prediction, \method learns to predict multiple future tokens simultaneously given past tokens in the ground-truth sequence. For backward reconstruction, we 
randomly corrupt tokens in the input sequence to create the intermediate states $\bm{y}^{(l)}$ in Eq.~\ref{eq:oa-llm}. Similar to BERT~\citep{DBLP:conf/naacl/DevlinCLT19}, we compute the loss only on the corrupted tokens. During training, we use the original data for prediction and the corrupted data for reconstruction. This decomposition disentangles the self-refinement capability from forward prediction, leveraging all data points to enhance sequence modeling.

\paragraph{Corruption Strategy.}
Our corruption and reconstruction process is a form of denoising autoencoding~\citep{DBLP:conf/icml/VincentLBM08} in language modeling. However, instead of representation learning, we aim to endow the model with the self-refinement capability to revise the generated content. Inspired by masked autoencoders~\citep{DBLP:conf/cvpr/HeCXLDG22}, we divide the output sequence into non-overlapping patches and randomly sample a subset for corruption. Each patch is a fragment of text containing one or multiple consecutive tokens in the sequence. Specifically, we corrupt a patch by either (\textbf{i}) replacing it with a random patch sampled from the current sequence or (\textbf{ii}) repeating the first token to replace the other tokens in the patch. This design draws on insight from \citet{ye2024physicslanguagemodels22} that model performance can be significantly improved by simply enhancing consistency across steps.

\subsection{Architecture: Target-Aware Query Representation for Self-Attention}\label{method:arc}
We build our framework by adapting the standard architecture of LLMs using decoder-only Transformers~\citep{DBLP:conf/nips/BrownMRSKDNSSAA20}. Unlike prior NAR works employing encoder--decoder architectures~\citep{DBLP:conf/emnlp/LeeMC18, DBLP:conf/icml/KasaiCGG20}, the conventional AR foundation predicts the same distribution given the current context regardless of the target token position. While this demonstrates advanced capabilities of sequence modeling and generation, the typical parameterization of next-token distribution constrains its generalizability to the order-agnostic objective in Eq.~\ref{eq:oa-llm}. 

Previous works on order-agnostic modeling have explored various ways to incorporate positional information,
including scaling up the dimensionality of the final projection layer~\citep{DBLP:conf/nips/SternSU18} and adding look-ahead tokens~\citep{DBLP:journals/corr/abs-2311-13581} or extra decoding heads~\citep{DBLP:conf/icml/CaiLGPLCD24,DBLP:conf/icml/GloeckleIRLS24}. Despite their promising performance, these methods introduce the overhead of additional self-attention network calls and new parameters for multi-position prediction. Instead, we propose a seamless adjustment without adding extra model parameters. Specifically, we apply a generalized Rotary Position Embedding (RoPE)~\citep{DBLP:journals/ijon/SuALPBL24} at the final layer of the decoder-only Transformers to integrate target-aware information into the query representations.

\paragraph{Target-Aware RoPE.}
RoPE encodes positional information into query and key representations, ensuring that their inner product inherently contains relative position information in self-attention:
\begin{equation}
    f(\bm{q}_m, m)^{\top} f(\bm{k}_n, n) = g(\bm{q}_m, \bm{k}_n, m - n)
    \label{eq:rope}
\end{equation}
where $f$ is the positional encoding function applied to the query and key embeddings at $m$-th and $n$-th positions, respectively, to produce the product $g$. Conventional RoPE integrates positional information of the current token to form the query representation. 
While this effectively enhances the position-aware representation of the input token in intermediate hidden states, it introduces inherent misalignment with the target token position when using the learned representation for output prediction. To avoid this problem, we propose Target-Aware RoPE (Figure~\ref{fig:model}), which modifies the positional encoding function at the final layer by considering the target token position in the query representation. Denoting $\mu$ as the position index of the target token within the context window to be predicted, we have:
\begin{equation}
    f(\bm{q}_m, \mu)^{\top} f(\bm{k}_n, n) = g(\bm{q}_m, \bm{k}_n, \mu - n), \quad \mu\in[m-k, m+k]
    \label{eq:tw-rope}
\end{equation}
The rationale behind this modification is that the position encoding in RoPE can adapt the representation of the current token to be tailored for the target position. This simple yet effective adjustment endows the model with the target-aware capability, allowing it to predict tokens at various positions without the overhead of additional entire network calls. 

\section{Sliding Blockwise Order-Agnostic Decoding}\label{sec:method-decode}
Leveraging the order-agnostic capabilities of \method, we propose Sliding Blockwise Order-Agnostic Decoding, a parallel decoding strategy to enable efficient iterative refinement.

High inference latency significantly hinders the broader application of AR-LLMs. Recent studies have tackled this bottleneck from various angles to accelerate inference. For instance, speculative decoding employs a smaller, faster draft model to propose multiple continuations, which the larger target model then verifies and accepts~\citep{DBLP:conf/icml/LeviathanKM23, DBLP:conf/asplos/MiaoOZCWZWZYSSC24}. Blockwise parallel decoding directly leverages the large model to generate multiple tokens simutaneously~\citep{DBLP:conf/nips/SternSU18, DBLP:conf/icml/CaiLGPLCD24}.
However, these studies increase memory consumption, which thus limits the scalability and impedes distributional deployment. 
Another promising line of work breaks the sequential dependency by adopting Jacobi decoding~\citep{DBLP:conf/acl/SantilliSPMMMR23, DBLP:conf/icml/FuBS024} for iterative refinement without architectural add-ons. \citet{DBLP:conf/icml/KouHHDZ24} propose consistency LLMs to further improve the performance of Jacobi decoding inspired by consistency models~\citep{DBLP:conf/icml/SongD0S23}. 

While these existing approaches improve inference efficiency, they rely on the conventional left-to-right AR foundation with monotonic dependencies. In this work, we leverage the order-agnostic nature of \method to perform backward sequence-level refinement and forward multi-token prediction simultaneously, significantly accelerating inference. At each step, we ensemble the output distributions based on multiple possible dependencies and construct a candidate set to fill a block of the output sequence. Furthermore, this process facilitates self-refinement by modifying previous generations at a higher-level horizon, enhancing output quality with advanced inference capabilities. Figure~\ref{fig:decoding} shows that this internal refinement process amends the duplicate ``marine'' to ``organism'', based on both forward and backward dependencies in the generated context. Next, we detail the ensemble strategy in decoding for candidate construction and verification.

\begin{algorithm*}[t]
    \small
    \caption{Sliding Blockwise Order-Agnostic Decoding}
    \label{alg:decode}
    \begin{algorithmic}[1]
        \STATE {\bfseries Input:} Order-agnostic generator $\pi_{\theta}$ and verifiers $v_{\theta}$ and $v_{\theta}^{\mathrm{CD}}$ based on OA-LLM $p_{\theta}$, prompt $\bm{x}$, decoding context window size $k$, decoding block size $b$, maximum output sequence length $T$.
        \STATE Initialize $t \leftarrow 0$, $\bm{y} \leftarrow \emptyset$. \COMMENT{Initialize the current length of the output sequence}
        \STATE Initialize $t_s \leftarrow 1$, $t_e \leftarrow \min{(k, b)}$. \COMMENT{Initialize the start and end indices of the block to predict and refine}
        \WHILE{$t_s < T$}
            \STATE Construct $\mathcal{Y}_{t_s:t_e} \leftarrow \left\{\{\tilde{y}_i\}_{i=t_s}^{t_e}, \tilde{y}_i\sim \pi_{\theta}(y_i\mid\bm{y}, \bm{x})\right\}$. \COMMENT{Collect candidates through tree construction}
            \STATE Select $\bm{y}_{t_s:t_e} \leftarrow \arg\max_{\tilde{\bm{y}}_{t_s:t_e}\sim\mathcal{Y}_{t_s:t_e}}\frac{1}{t_e - t_s + 1}\sum_{i=t_s}^{t_e}\left(v_{\theta}(\tilde{y}_i\mid\bm{y}, \bm{x}) + v_{\theta}^{\mathrm{CD}}(\tilde{y}_i\mid\bm{y}, \bm{x})\right)$. \COMMENT{Verify}
            \STATE Update $\bm{y} \leftarrow \mathrm{concat}(\bm{y}_{< t_s}, \bm{y}_{t_s:t_e})$.
            \STATE Set $t \leftarrow t_e$.
            \FOR{$i=t_s$ {\bfseries to} $t_e$}
                \STATE Sample $r \sim U[0, 1]$ from a uniform distribution
                \IF{$r < c(y_i\mid \bm{y}, \bm{x})$}
                    \STATE Set $t_s \leftarrow t_s + 1$. \COMMENT{Slide the decoding block based on rejection sampling}
                    \IF{$y_i$ $==$ \texttt{[EOS]}}
                        \STATE Exit while loop.
                    \ENDIF
                \ELSE
                    \STATE Exit for loop.
                \ENDIF
            \ENDFOR
            \STATE Set $t_e \leftarrow \min{(t_s + b - 1, t + k)}$.
        \ENDWHILE
        \STATE {\bfseries Output:} $\bm{y}$
    \end{algorithmic}
\end{algorithm*}

\paragraph{Prediction.}
Given a set of possible distributions $\left\{p_{\theta}(y_t\mid \bm{y}_{\leq i}, \bm{x})\right\}_{i=t-k}^{t+k}$ for the $t$-th token in the output sequence, we obtain the ensemble distribution via model arithmetic~\citep{DBLP:conf/iclr/Dekoninck0BV24}. Specifically, we apply different weights to the distributions to prioritize the more accurate dependencies, with distributions based on more qualified content generally leading to better generations:
\begin{equation}
    \vspace{-.02in}
    \pi_{\theta}(y_t) = \mathrm{softmax}\left(\frac{1}{\sum_{i=t-k}^{t+k}\omega_{t - i}(\bm{y}_{\leq i}, \bm{x})}\sum_{i=t-k}^{t+k}\omega_{t - i}(\bm{y}_{\leq i}, \bm{x})\log p_{\theta}(y_t\mid\bm{y}_{\leq i}, \bm{x})\right)
    \label{eq:oa-decode}
\end{equation}
The weight function $\omega_{t-i}(\bm{y}_{\leq i}, \bm{x}) = \lambda_{t-i}\cdot c(\bm{y}_{\leq i}\mid\bm{x})$ is determined by the relative distance and direction of the dependency, as well as the confidence of the generated context $\bm{y}_{\leq i}$. Here, the factor $\lambda_{t-i}\in[0, 1]$ only depends on the relative position of the target token, decaying for longer dependencies. Using order-agnostic modeling, we calculate the confidence score $c$ by gathering the predicted probabilities based on different dependencies, which we obtain in the verification stage. Generally, backward reconstruction and next-token prediction based on iteratively refined content will be associated with higher weights. 
See Section~\ref{sec:further} for a detailed comparison among different dependencies. 
In practice, some of the distributions in Eq.~\ref{eq:oa-decode} may not be available for all tokens at each step. We calculate the ensemble utilizing available dependencies within the context window.

\paragraph{Verification.}
Following \citet{DBLP:conf/icml/CaiLGPLCD24}, we employ tree attention\footnote{To balance exploitation and exploration in tree construction, we select nodes according to the estimated accuracy of each token. Detailed considerations of candidate selection can be found in Appendix~\ref{app:decode}.} to select from multiple candidates sampled from the ensemble distribution $\pi_{\theta}$. Each candidate is a combination of tokens used to fill the sliding block. Unlike previous works that only adopt the original next-token probability for verification, we also incorporate the backward reconstruction probabilities to leverage the refinement ability of \method. The verification score can thereby be formulated as follows:
\begin{equation}
    \vspace{-.02in}
    v_{\theta}(y_t) = \frac{1}{\sum_{i=t-1}^{t+k}\lambda_{t - i}}\sum_{i=t-1}^{t+k}\lambda_{t - i}\log p_{\theta}(y_t\mid\bm{y}_{\leq i}, \bm{x})
    \label{eq:oa-verify}
\end{equation}
Here, we only consider the next-token and backward predictions for the verification score calculation. This scheme can be further enhanced by introducing a contrastive objective~\citep{DBLP:conf/acl/LiHFLEHZL23} that penalizes the possible failure cases in forward multi-token prediction:
\begin{equation}
    \vspace{-.02in}
    v_{\theta}^{\mathrm{CD}}(y_t) = \max\left(0,\log{p_{\theta}(y_t\mid\bm{y}_{\leq t-1}, \bm{x})} - \frac{1}{\sum_{i=t-k}^{t-2}\lambda^{\prime}_{t-i}}\lambda^{\prime}_{t-i}\log{p_{\theta}(y_t\mid \bm{y}_{\leq i}, \bm{x})}\right)
    \label{eq:oa-verify-cd}
\end{equation}
where $\lambda^{\prime}_{t-i}=1/\lambda_{t-i}$ to apply a higher penalty to predictions based on longer dependencies. Combining $v_{\theta}$ with $v_{\theta}^{\mathrm{CD}}$, we keep the candidate of the highest average score. We allow several refinement iterations for each position within a sliding block to enhance the generation quality. 
Specifically, we propose an ensemble rejection sampling scheme to determine the sliding step size through majority voting across multiple dependencies, where we accept each token with the probability:
\begin{equation}
    \vspace{-.02in}
    c(y_t\mid\bm{y}_{\leq t+k},\bm{x}) = \frac{1}{k + 2}\sum_{i=t-1}^{t+k}\mathbb{1}_{p_{\theta}(y_t\mid\bm{y}_{\leq i},\bm{x}) > \min\left(\epsilon, \epsilon\exp\left(-H\left(p_{\theta}(\cdot\mid\bm{y}_{\leq i},\bm{x})\right)\right)\right)}
    \label{eq:conf}
\end{equation}
where $H(\cdot)$ is the entropy and $\epsilon$ is a fixed threshold to reject low-probability predictions. This acceptance scheme is inspired by truncation sampling~\citep{DBLP:conf/emnlp/HewittML22,DBLP:conf/icml/CaiLGPLCD24} to choose candidates that are more likely to be sampled from the reference distributions. The sliding step size for each step is set to the length of the longest accepted prefix of the current block. We detail the sliding decoding procedure in Algorithm~\ref{alg:decode}.

\section{Experiments}\label{sec:exp}
In this section, we demonstrate the efficiency and breadth of \method regarding the quality--speed trade-offs across arithmetic, logical reasoning, and code generation. 

\paragraph{Datasets.}
For arithmetic reasoning, we train \method on MetaMathQA ($395$K)~\citep{DBLP:conf/iclr/YuJSYLZKLWL24} and evaluate it using GSM8K~\citep{DBLP:journals/corr/abs-2110-14168} on grade school math word problems and MATH~\citep{DBLP:conf/nips/HendrycksBKABTS21} of challenging competition mathematics problems. For logical reasoning, we filter LogiCoT~\citep{DBLP:conf/emnlp/LiuTCZZ023} with deduplication and reformulation, obtaining $313$K training samples. We assess logical reasoning performance with multiple-choice reading comprehension tasks that test interpretation and decision-making skills: LogiQA~\citep{DBLP:journals/taslp/LiuLCTDZZ23}, based on the Chinese Civil Service Examination, and ReClor~\citep{DBLP:conf/iclr/YuJDF20}, sourced from Law School Admission Council exams. For code generation, we train on Magicoder-Eval-Instruct-110K~\citep{DBLP:journals/corr/abs-2312-02120} and evaluate using programming tasks from HumanEval~\citep{DBLP:journals/corr/abs-2107-03374}.

\paragraph{Experimental Protocol.}
To address the discrepancy between the pre-trained model based on next-token dependency and the target order-agnostic model, we adopt a two-stage training strategy~\citep{DBLP:conf/iclr/KumarRJ0L22} to progressively enhance order-agnostic modeling. We begin with a domain-specific supervised fine-tuned (SFT) model for each target task. In the first stage, we perform order-agnostic training exclusively on the last target-aware layer, while freezing the other layers to preserve the output quality. In the second stage, following \citet{DBLP:conf/icml/CaiLGPLCD24}, we train the entire model by focusing on the previously frozen layers first and then unlocking the last layer to train together.  We use Mistral-7B-v0.3
as the base models for reasoning and code generation tasks, respectively. During inference, we explore the effect of the verification stage and ablate the values of decoding context window size and block size. We detail our hyperparameter settings in Section~\ref{sec:ablate} and Appendix~\ref{app:para}.

\subsection{Main Results}\label{exp:rst}
We compare our order-agnostic decoding approach (Section~\ref{sec:method-decode}) with its next-token counterparts across three tasks. We also show the quality--speed trade-offs in by ablating the decoding settings. 
\begin{table}[t]
    \small
    \caption{Result comparison of performance (accuracy $\%$) and speed (accepted tokens per second) on arithmetic reasoning tasks. We compare against the conventional autoregressive greedy decoding approach as our next-token prediction baseline (NT). ``verifier'' and ``multi-forward'' represent the verification stage and multiple forward token prediction in inference.}
    \label{tab:arithmetic}
    \begin{center}
    \vskip -.15in
    \begin{tabular}{lccccccc}
    \toprule
    \multirow{2}{*}{\bf Approach}  & \multicolumn{3}{c}{\bf GSM8K} & & \multicolumn{3}{c}{\bf MATH} \\
    \cmidrule{2-4} \cmidrule{6-8} 
    & Accu. & Speed & Speedup & & Accu. & Speed & Speedup \\
    \midrule
    NT &  $74.1$ & $39.7$ & $1.0\times$ & & $21.8$ & $38.7$ & $1.0\times$ \\
    \midrule
    Ours & \deltavalup{75.3}{1.2} & $43.4$ & $1.1\times$ && \deltavalup{22.7}{0.9} &  $44.4$ & $1.1\times$ \\
    Ours $_\textrm{w/o verifier}$ & \deltavaldown{72.4}{1.7} & $156.8$ & $\mathbf{3.9}\times$ && \deltavaldown{20.0}{1.8} & $139.7$ & $\mathbf{3.6}\times$ \\
    Ours $_\textrm{w/o multi-forward}$ & \deltavalup{\mathbf{78.7}}{4.6} & $14.9$ & $-$ & & \deltavalup{\mathbf{24.3}}{2.5} & $11.5$ & $-$ \\
    \bottomrule
    \end{tabular}
    \end{center}
    \vskip -.1in
\end{table}

\paragraph{Arithmetic Reasoning.}
As shown in Table~\ref{tab:arithmetic}, \method enhances the effectiveness and efficiency through different mechanisms in order-agnostic generation. 
Using both verification and multiple forward token prediction in decoding, \method surpasses the corresponding next-token baseline with comparable inference-time cost. 
Furthermore, by trading inference speed with iterative generation and verification through backward refinement, we observe a substantial improvement in accuracy from $74.1$\% to $78.7$ and $21.8\%$ to $24.3\%$, on GSM8K and MATH, respectively. When skipping the verification stage for quality control, our approach significantly speeds up the decoding process up to $3.9\times$. This demonstrates the flexibility of \method in enhancing both the generation quality and inference speed in mathematical reasoning.

\begin{table}[t]
    \small
    \caption{Result comparison of performance and speed on logical reasoning tasks.}
    \label{tab:logical}
    \begin{center}
    \vskip -.15in
    \begin{tabular}{lccccccc}
    \toprule
    \multirow{2}{*}{\bf Approach}  & \multicolumn{3}{c}{\bf LogiQA} & & \multicolumn{3}{c}{\bf ReClor} \\
    \cmidrule{2-4} \cmidrule{6-8}
    & Accu. & Speed & Speedup & & Accu. & Speed & Speedup \\
    \midrule
    NT &  $55.1$ & $33.6$& $1.0\times$ & & $63.2$ & $33.2$ & $1.0\times$ \\
    \midrule
    Ours & \deltavalup{58.2}{3.1} & $62.1$ & $1.8\times$ && \deltavaldown{62.7}{0.5} & $38.2$ & $1.2\times$  \\
    Ours $_\textrm{w/o verifier}$ & \deltavalup{55.7}{0.6} & $99.1$ & $\mathbf{2.9}\times$ && \deltavaldown{61.6}{1.6} & $72.0$ & $\mathbf{2.2}\times$ \\
    Ours $_\textrm{w/o multi-forward}$ & \deltavalup{\mathbf{59.1}}{4.0} & $8.9$ & $-$ && \deltavalup{\mathbf{64.7}}{1.5} & $11.3$ & $-$ \\
    \bottomrule
    \end{tabular}
    \end{center}
    \vskip -.1in
\end{table}
\begin{figure*}[t]
    \centering
    \begin{minipage}[c]{0.48\textwidth}
        \centering
        \captionsetup{type=table, position=top}
        \caption{Result comparison of pass rates and speed on code generation.}
        \label{tab:he}
        \small
        \begin{tabular}{lccc}
            \toprule
            \multirow{2}{*}{\bf Approach}  & \multicolumn{3}{c}{\bf HumanEval} \\
            \cmidrule{2-4}
            & Pass@$1$ & Speed & Speedup \\
            \midrule
            NT &  $\mathbf{64.6}$ & $42.2$& $1.0\times$ \\
            \midrule
            Ours & \deltavaldown{13.0}{51.6} & $45.8$ & $1.1\times$  \\
            Ours $_\textrm{w/o verifier}$ & \deltavaldown{6.5}{58.1} & $119.0$ & $\mathbf{2.8}\times$ \\
            Ours $_\textrm{w/o multi-forward}$ & \deltavaldown{61.6}{3.0} & $28.8$ & $-$ \\
            \bottomrule
        \end{tabular}
    \end{minipage}
    \hfill
    \begin{minipage}[c]{0.48\textwidth}
    \centering
    \includegraphics[width=\textwidth]{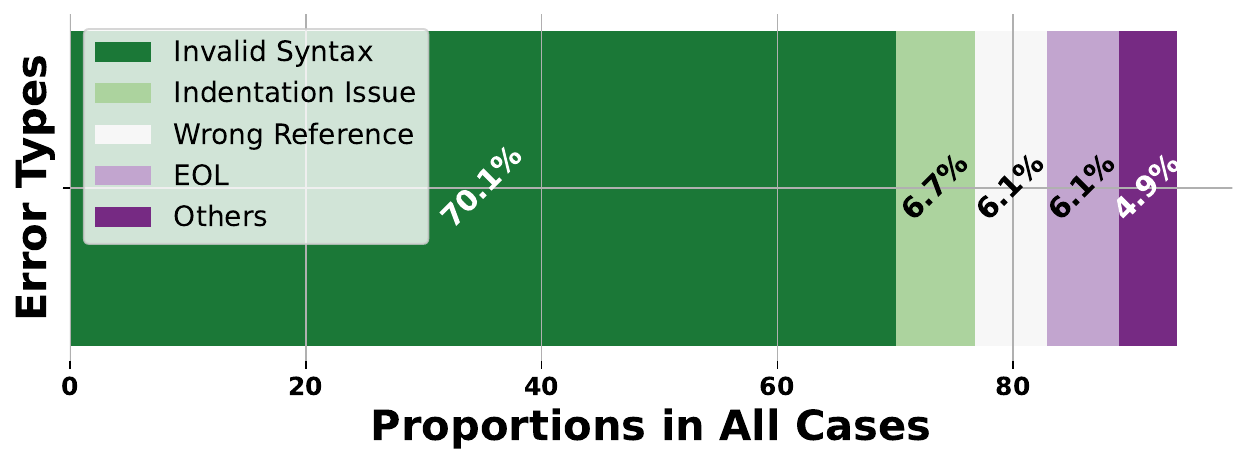}
    \caption{Meso-analysis of error cases in code generation (Ours $_\textrm{w/o verifier}$) on HumanEval. The primary failure cases come from syntax errors.}
    \label{fig:he-error}
    \end{minipage}
    \vskip -.1in
\end{figure*}

\paragraph{Logical Reasoning.}
Table~\ref{tab:logical} compares the performance and generation speed of model outputs under different decoding settings on logical reasoning tasks. Similarly, \method improves the reasoning performance by augmenting next-token prediction exclusively with backward refinement. However, we observe a discrepancy in the performance improvements on LogiQA and ReClor with absolute increases of $4.0\%$ and $1.5\%$ in corresponding accuracies. We attribute this gap to the imbalanced proportions of the two tasks in our SFT data from LogiCoT~\citep{DBLP:conf/emnlp/LiuTCZZ023}. This also implies the importance of high-quality data selection to boost the effect of order-agnostic training to model different dependencies related to the target tasks.

\paragraph{Code Generation.}
Results on code generation, however, show an opposite effect of order-agnostic modeling on performance. In Table~\ref{tab:he}, we observe substantial performance drops across different decoding settings using \method. For example, without verification, the pass rate on HumanEval decreases to $6.5\%$ from $64.6\%$ of next-token prediction. This gap remains to be large when applying verification for quality control. Error analysis in Figure~\ref{fig:he-error} indicates that the major cause of this drop comes from the erroneous syntax, where the primary error type, \textit{Invalid Syntax}, accounts for $70.1\%$ of all samples. 
To mitigate this issue, we can turn off the mechanism of forward multi-token prediction and increase the threshold $\epsilon$ in Eq.~\ref{eq:conf} to reject tokens with low confidence scores. For example, with $\epsilon=0.5$, \method achieves a comparable pass rate of $61.6\%$ compared to $64.6\%$ of the baseline. The absolute decrease of $3.0\%$ indicates the deficiency of \method in producing incoherent content, showing the importance of specific designs for tasks requiring strict textual formats.

\subsection{Ablation Studies}
\label{sec:ablate}
In this section, we analyze the core designs of \method to enable efficient iterative refinement. We probe the effect of different training and decoding hyperparameters.

\begin{figure}[t]
    \centering
    \begin{subfigure}[b]{0.48\textwidth}
        \centering
        \includegraphics[width=\textwidth]{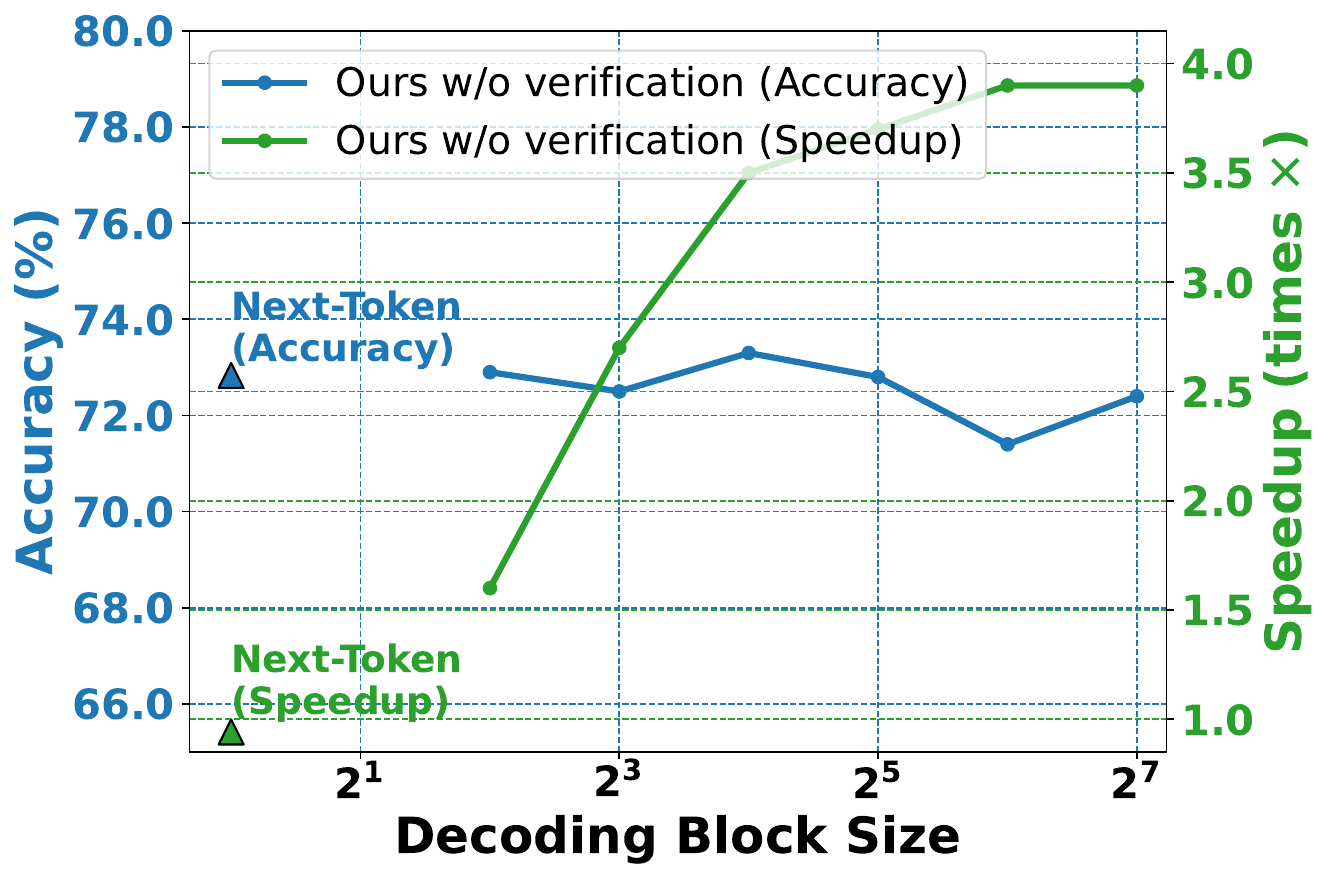}
        \caption{Block size for acceleration.}
        \label{fig:b}
    \end{subfigure}
    \quad
    \begin{subfigure}[b]{0.48\textwidth}
        \centering
        \includegraphics[width=\textwidth]{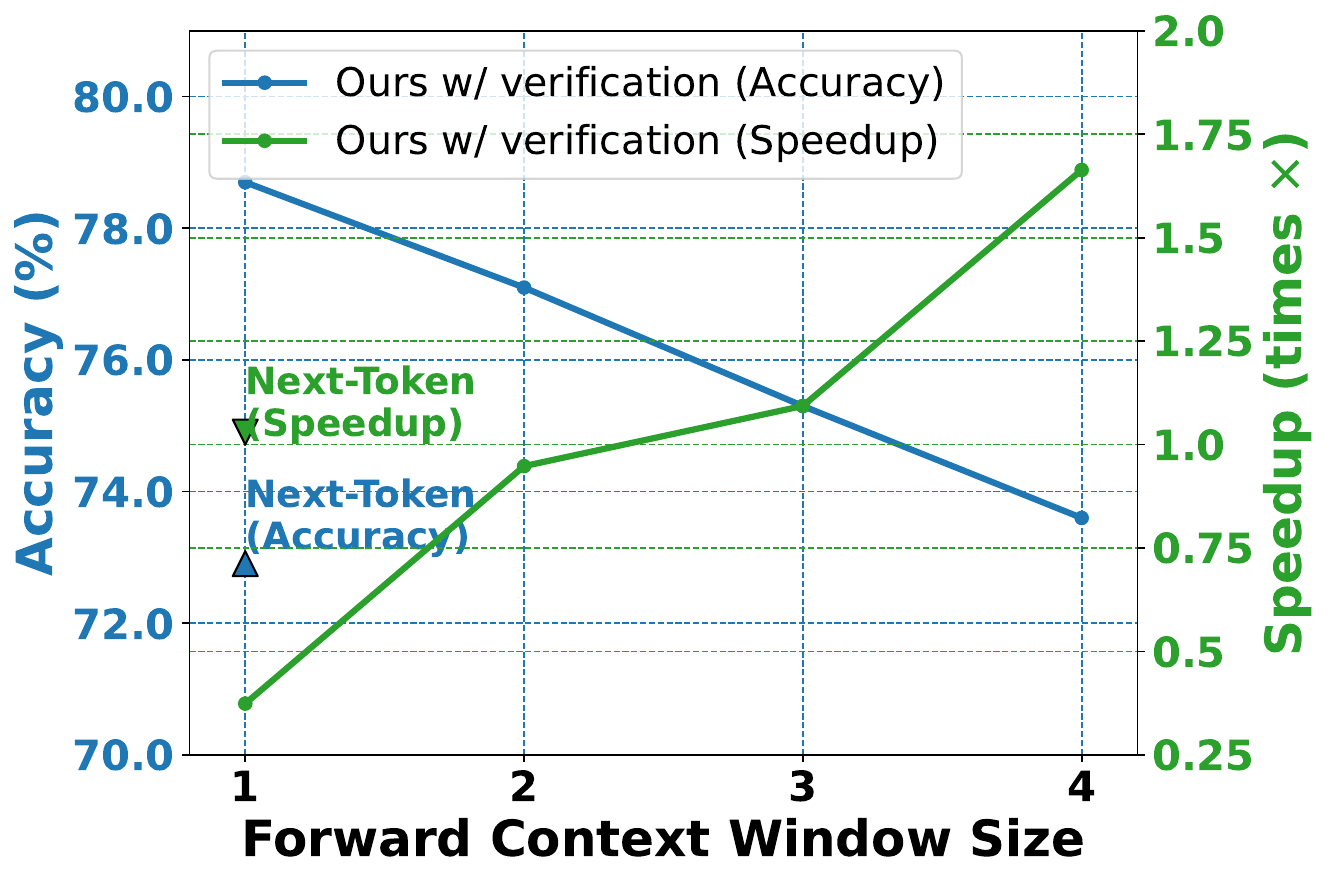}
        \caption{Quality--speed trade-off.}
        \label{fig:k}
    \end{subfigure}
    \caption{Quality--speed trade-offs on GSM8K. Generally, \method accelerates inference with larger decoding block size $b$ and forward context window size $k$. }
    \label{fig:trade-off}
    \vskip -.1in
\end{figure}
\begin{figure}[t]
    \centering
    \begin{subfigure}[b]{0.48\textwidth}
        \centering
        \includegraphics[width=\textwidth]{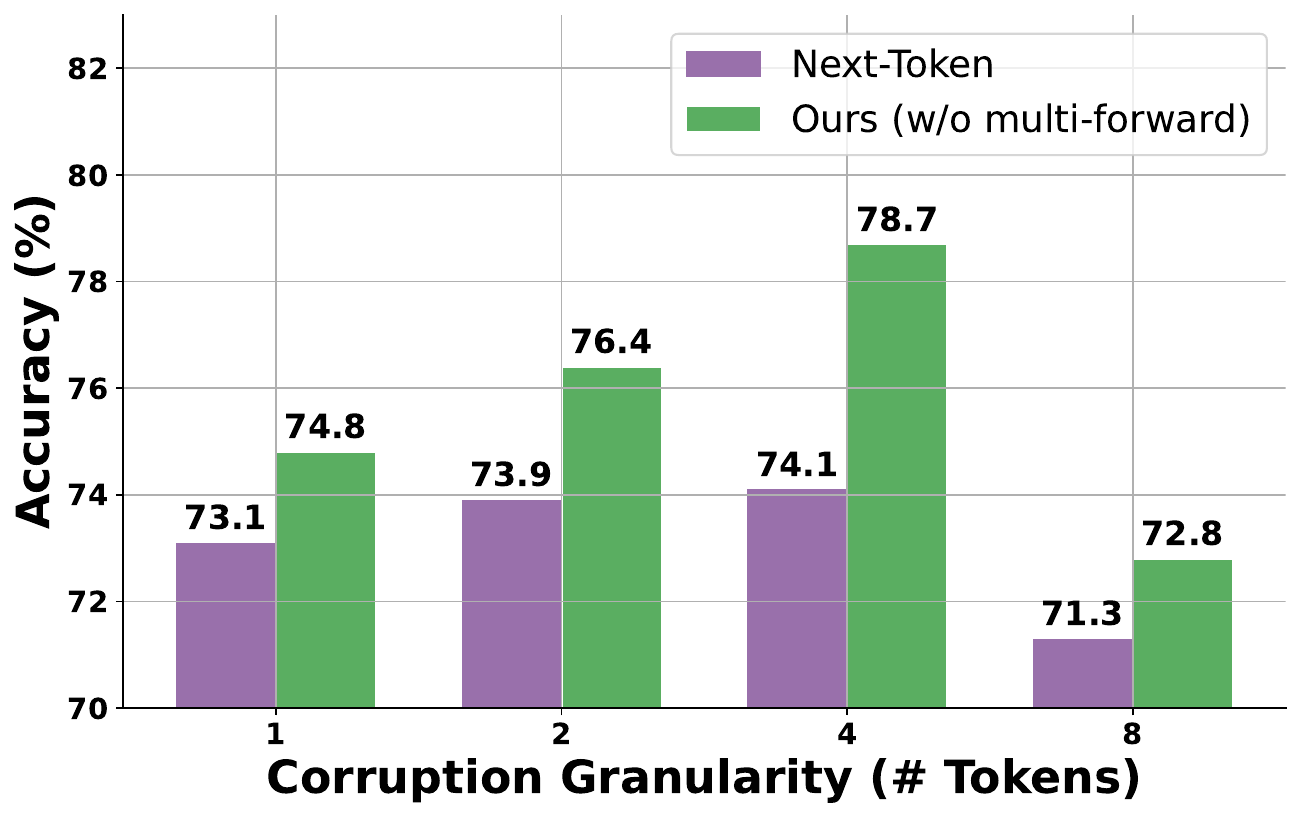}
        \caption{Effect of corruption granularity.}
        \label{fig:corrupt-g}
    \end{subfigure}
    \quad
    \begin{subfigure}[b]{0.48\textwidth}
        \centering
        \includegraphics[width=\textwidth]{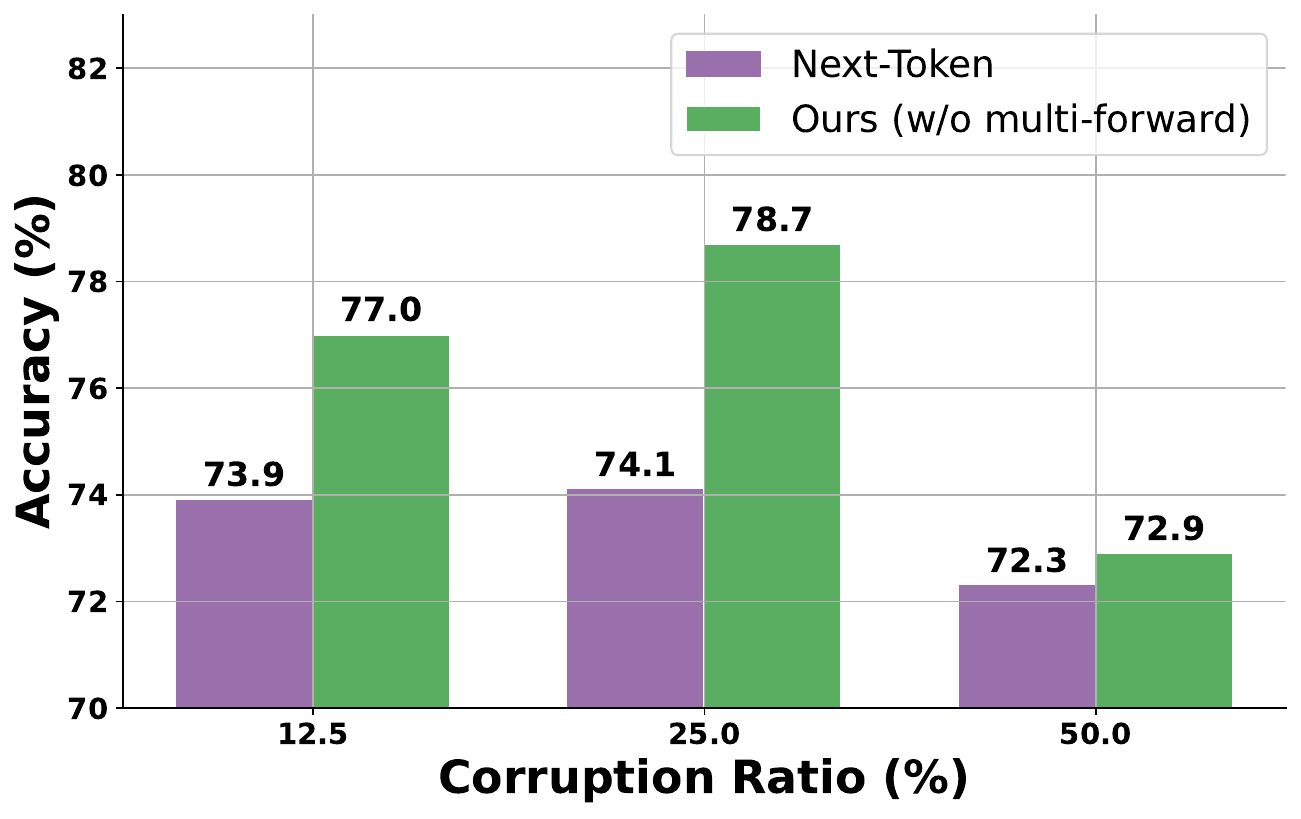}
        \caption{Effect of corruption ratio.}
        \label{fig:corrupt-r}
    \end{subfigure}
    \caption{Ablation on the effects of corruption granularity and ratios in training for backward reconstruction. We probe the variation in model improvements from backward dependencies.}
    \label{fig:corrupt}
    \vskip -.1in
\end{figure}

\begin{figure}[t]
    \centering
    \begin{subfigure}[b]{0.48\textwidth}
        \centering
        \includegraphics[width=\textwidth]{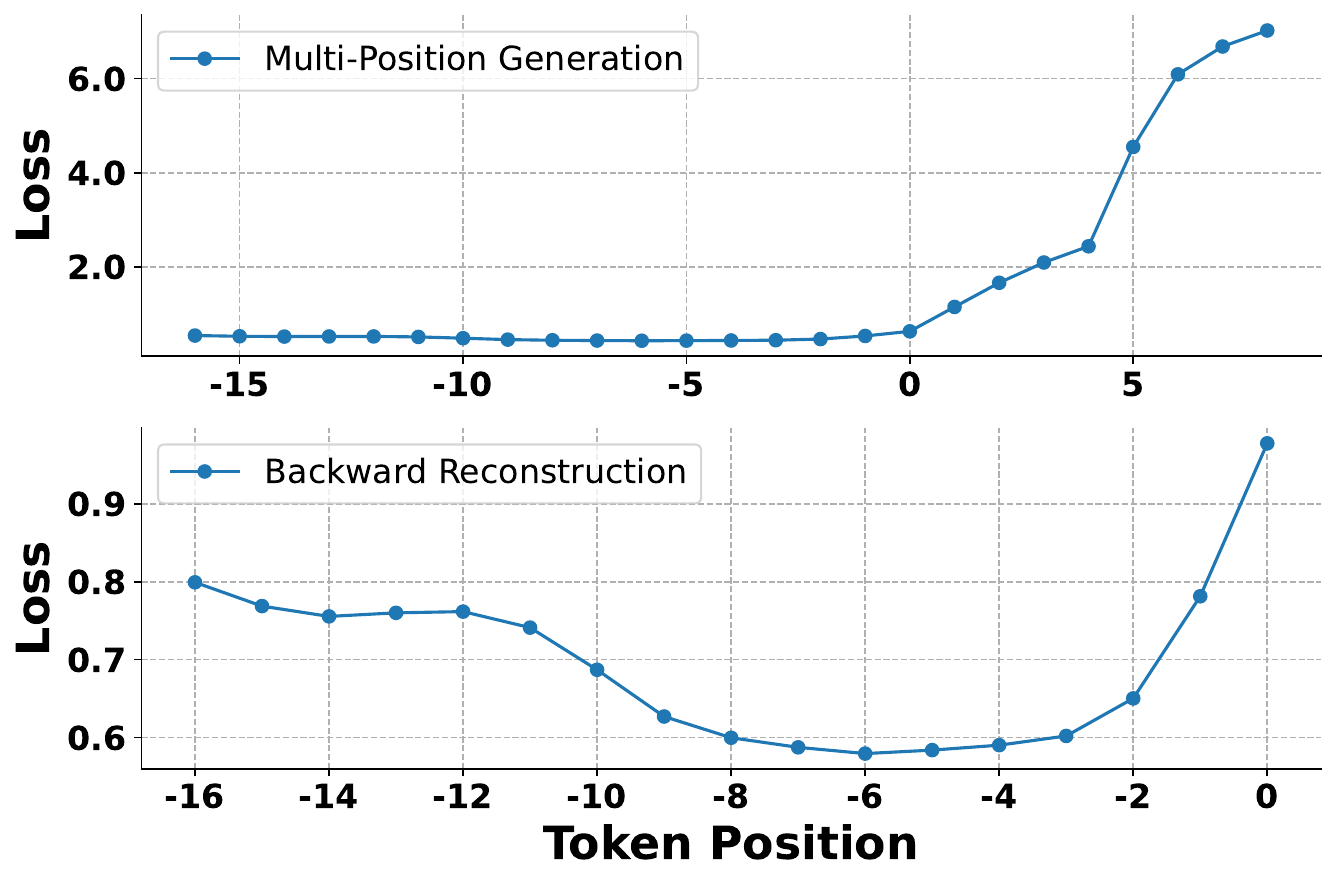}
        \caption{Token-wise loss.}
        \label{fig:loss}
    \end{subfigure}
    \quad
    \begin{subfigure}[b]{0.48\textwidth}
        \centering
        \includegraphics[width=\textwidth]{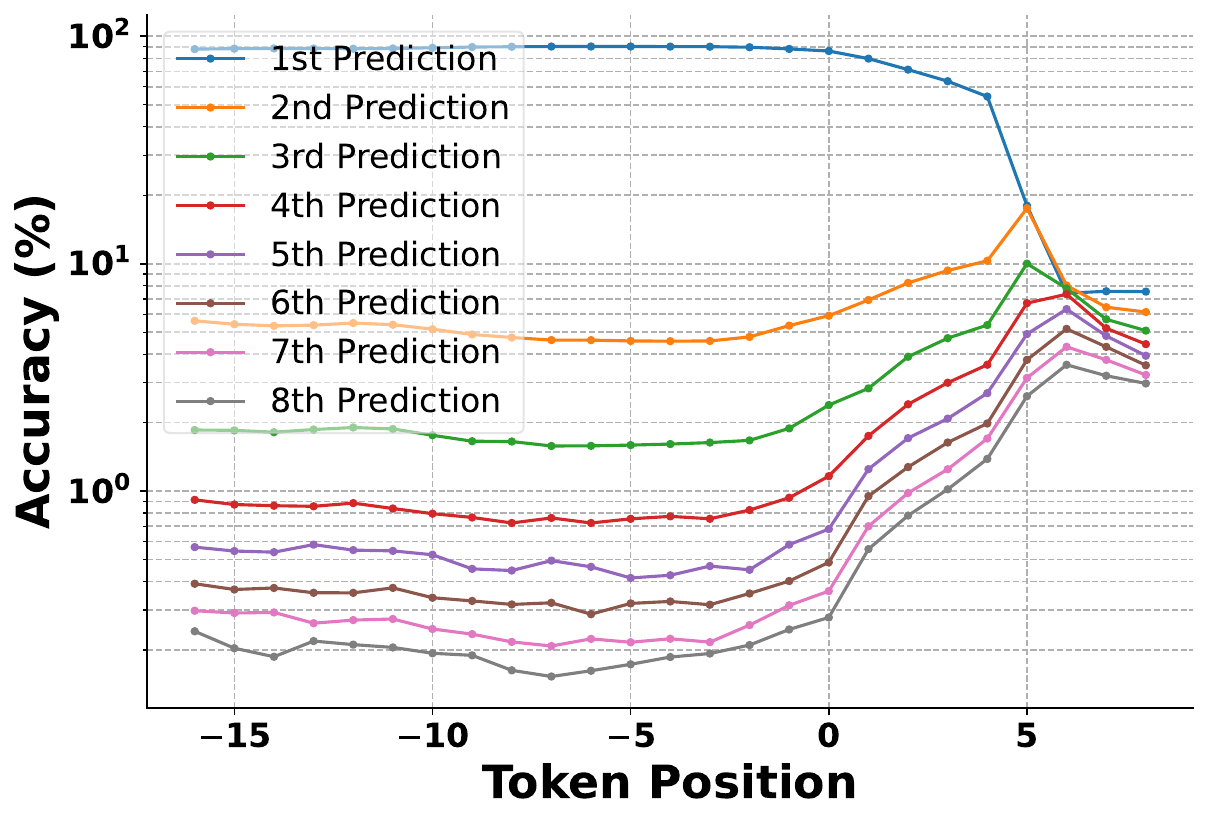}
        \caption{Token-wise accuracy of top-$8$ predictions.}
        \label{fig:accu}
    \end{subfigure}
    \caption{Token-wise losses and accuracies corresponding to different dependencies.}
    \label{fig:accu-loss}
    \vskip -.1in
\end{figure}

\paragraph{Backward Refinement Improves Generation Quality.}
Figure~\ref{fig:it} shows how the performance and inference cost scale with  iterative refinement. Note that even without backward dependencies in prediction, \method can still perform backward refinement using the next-token prediction. In this case, we examine the effect of backward dependencies with different context window sizes. Notably, the performance of iterative refinement scales faster than the inference cost as the iteration time increases. Furthermore, leveraging backward dependencies, \method reaches a higher plateau of performance compared to refining with forward dependencies only. However, the fast saturation of performance improvement with larger refinement times indicates a relatively low upper bound of the enhancement brought by backward reconstruction. We extensively discuss this problem attributed to the discrepancy between pre-training and fine-tuning objectives in Appendix~\ref{app:ext}. 

\paragraph{Quality--Speed Trade-off in Inference.}
In Section~\ref{exp:rst}, we demonstrate the quality--speed trade-off by ablating the employment of verification and forward multi-token prediction. We now provide a detailed analysis of the decoding hyperparameters to show this trade-off. We consider the block size $b$ and the forward context window size $k$, two variables closely related to the inference speed and quality. For block size, we probe its effect in reducing inference time in the verification-free case to maximize the speedup rate. Figure~\ref{fig:b} shows that we can push the speedup boundary toward the corresponding upper bound of $k$ with large $b$. For example, given $k\!=\!4$, we approach the maximum speedup rate $4\times$ with large block sizes such as $b\!=\!64$ and $128$. Notably, leveraging the backward refinement capability of \method, this block size-driven acceleration process retains the generation quality at the same level, illustrating the balance of efficiency and effectiveness of our acceleration mechanism. 
For forward context window size, we adopt the two-stage prediction--verification setting to explore the quality improvement boundary of the iterative refinement mechanism. Figure~\ref{fig:k} shows trends of performance drop and inference speedup when increasing $k$. We explain this trade-off as a reflection of the decreasing precision in predicting future tokens of longer dependencies. We conduct further analysis on the modeling ability for different dependencies in Section~\ref{sec:further}.

\paragraph{Learning from Corruption Enhances Refinement Capability.}
One core design in \method is the denoising process to enable iterative refinement, where the corruption strategy is crucial for controlling data quality and model performance. In Figure~\ref{fig:corrupt}, we analyze the variation in performance improvements from backward refinement (w/ multi-forward) when applying corruption with different granularity or ratios. Given a backward context window size $k=8$, we observe a more significant improvement when applying corruption on longer pieces of text. For example, \method achieves an absolute increase of $4.6\%$ with granularity $4$, compared to $1.7\%$ under token-level corruption. However, as the corrupted context gets longer, the model's capability to learn from mistakes may also degrade. One possible reason for this performance drop is the difficulty and inconsistency in simultaneous multi-token regeneration, as reconstructing more tokens brings higher uncertainty and noise. This indicates the importance of using a reasonable corruption granularity to obtain data of good quality and maintain training stability. 
Likewise, we see a similar trend when the corruption ratio varies. Specifically, a high corruption ratio such as $0.5$ can damage the semantic meaning of the context, leading to a performance drop in both our and baseline approaches. Nevertheless, we can still benefit when increasing the corruption ratio within a reasonably lower range, such as $0.125$ to $0.25$, to enhance the reconstruction process.

\subsection{Further Analysis}
~\label{sec:further}
We analyze \method{}'s capability to model different dependencies, and the potential computation overhead from order-agnostic modeling. 

\paragraph{How does \method model order-agnostic dependencies?}
We compare the model capabilities across different positions using token-wise losses and accuracies in Figure~\ref{fig:accu-loss}. Generally, \method performs better on backward reconstruction than forward prediction, as shown in the lower losses and higher accuracies on backward dependencies. Notably, we see better generalizability of backward reconstruction. For example, given the backward context window size $k=8$ and forward context window size $k=4$, we find that the loss and accuracy of backward reconstruction with dependencies longer than the training context window size, such as positions $|\!-\!9|\!>\!|\!-\!8|$, are also at the same level as other backward dependencies. Differently, we observe a dramatic increase in loss and a drop in accuracy from positions $4$ to $5$ on longer dependencies in forward prediction. This explains how backward refinement benefits from more information in sequence-level generation to improve performance. We observe decreased performance for forward prediction as the dependency gets longer, especially when it exceeds the forward context window size in training. However, we can mitigate this issue by aggregating multiple predictions for each position. As shown in Figure~\ref{fig:accu}, while forward positions with longer dependencies obtain lower accuracies on tghe first prediction, the accumulated accuracies of their non-first predictions are generally higher than those from other dependencies. This illustrates how \method can benefit from the tree construction and verification stage in decoding (Section~\ref{sec:method-decode}) by considering multiple candidates for each position. 

\paragraph{Computation Overhead.}
One concern regarding order-agnostic modeling is the potential computation overhead to accommodate more dependencies in the context windows. As target-aware RoPE is only applied on the last layer, this overhead scales relatively slower as we increase the number of positions to predict. For example, with forward and backward context window sizes each set as $k=4$, each forward pass of \method costs $5.48$ TFLOPS, compared with $2.81$ TFLOPS of next-token prediction. In other words, \method predicts $8\times$ number of tokens with less than $2\times$ overhead in computational cost. This indicates the efficiency of \method in leveraging available computation resources to accelerate and enhance inference. Furthermore, we can adjust the forward and backward context window sizes to determine the number of tokens to predict in parallel, demonstrating the flexibility and generalizability of \method with target-aware RoPE.

\paragraph{Effect of Two-Stage Training.}
As discussed in Section~\ref{sec:ablate}, a high corruption ratio can cause a collapse in model performance as the noisy data contains corrupted information in a format that the model has not seen in pretraining. Furthermore, we are also faced with the order-agnostic training tax to endow an AR-based LLM with denoising and multi-token prediction abilities. In this section, we elaborate on the two-stage training we designed to mitigate this issue. Following \citet{DBLP:conf/icml/CaiLGPLCD24}, we first tune the last layer where we apply target-aware RoPE. This adapts the previous parameterization on next-token prediction to target-aware multi-position prediction. Due to the discrepancy of training objectives in pretraining and fine-tuning, full fine-tuning is still essential to ensure better performance on multi-token prediction. To stabilize the training process, we then freeze the last layer and gradually unlock it through the second training stage of full fine-tuning. Empirically, we find this strategy effective for stabilizing the autoregressive loss changes in forward prediction. However, we observe an order-agnostic training tax where the next-token prediction performance drops from $77.0\%$ to $76.5\%$ and then $74.1\%$ after the first and second stages, respectively. This performance degradation possibly comes from two aspects: the difference in training objectives and the incorporation of corrupted data in fine-tuning. We leave it to future work to further explore the effect of applying our order-agnostic framework to the pretraining stage.
\section{Discussion and Conclusion}
\label{sec:conclusion}
By unifying denoising with context-wise order-agnostic language modeling and introducing target-aware positional encoding, \method incorporates iterative refinement directly into the language generation process while keeping inference costs low. This approach offers a promising direction for developing more efficient and capable large language models by effectively capturing local dependencies within context windows and reducing inference latency. 

The effectiveness and efficiency of \method underscores the promise of order-agnostic strategies as a generalized architecture to facilitate generative language modeling and text generation. Specifically, it suggests new opportunities to unify: (\textbf{i}) the sequence modeling and varying-length generation abilities of autoregressive modeling, (\textbf{ii}) the multi-dependency modeling and multi-token prediction mechanisms in order-agnostic modeling, and (\textbf{iii}) the efficient way of iterative refinement in denoising techniques. We hope our work will motivate future research to explore order-agnostic modeling and denoising in various tasks and other domains beyond sequence modeling.

\section*{Limitations}
This work proposes an approach to integrate iterative denoising with order-agnostic language modeling to enhance both the effectiveness and efficiency of LLM inference. While it offers a promising paradigm for mitigating issues related to monotonic dependencies and inference latency in conventional autoregressive models, several directions remain for further exploration, including designing corruption and decoding strategies to tailor the model to specific tasks, optimizing the training process to overcome the order-agnostic training tax, and probing the generalizability and scalability of \method across different context sizes, model scales, and tasks.

Specifically, order-agnostic language modeling can struggle with tasks that demand specific output formats or syntax due to inconsistencies in the multi-token predictions. This indicates the importance of a task-specific design of the acceptance scheme in order-agnostic decoding. For instance, the performance of the verification policy in Eq.~\ref{eq:oa-verify} may vary by language and domain. Additionally, applying semantic-aware weights to different dependencies could further enhance task-specific features in the generated outputs. Future work can further explore the potential of incorporating different evaluation heuristics to guide the inference process. 

Furthermore, incorporating corrupted data may introduce discrepancies between training- and inference-time objectives. For example, our experiments explore rule-based context-wise corruption strategies to create noisy data.  Future work could focus on diversifying the types of corruption and scaling the difficulty level and proportion to better understand their impacts on model capabilities.

Finally, due to the computation constraint, we explore the model capabilities in order-agnostic modeling with fixed context window sizes during the SFT stage only. Future work may investigate the effect of scaling context window sizes in both forward and backward directions. Moreover, increasing the context window sizes may exacerbate the discrepancy between autoregressive pre-training and order-agnostic fine-tuning. We thus anticipate future work to extend \method to the pre-training stage to further enhance model capabilities. 

\section*{Potential Broader Impact}
Compared to conventional autoregressive modeling, \method leverages multi-token prediction and reconstruction to backtrack and iteratively refine past generations. This strategy mirrors the human decision-making process in real-world task completion. We anticipate \method to inspire the community to design more efficient and effective frameworks to enhance interpretability and alignment with the reasoning and planning process of humans.

\begingroup
\small
\bibliography{iclr2025_conference}
\bibliographystyle{iclr2025_conference}
\endgroup

\clearpage
\appendix
\section{Conceptual Comparison among Model Architectures}
\label{app:comp}
We consider the properties an ideal architecture should have as follows:
\begin{itemize}
    \item \textbf{VL}: varying-length generation
    \item \textbf{BT}: backtrack / look-ahead
    \item \textbf{MV}: multi-variable generation
    \item \textbf{MD}: multi-dependency (inter-sample connection) modeling
    \item \textbf{FS}: fitting feasibility
    \item \textbf{EF}: inference efficiency
    \item \textbf{IT}: mechanism of iterative refinement
\end{itemize}
\begin{table}[h]
\caption{Conceptual comparison regarding desired features across different architectures.}
\label{tab:arc-compare}
\vspace{-0.1in}
\begin{center}
    \begin{tabular}{lccccccc}
        \toprule
        \multicolumn{1}{c}{\bf Architectures}  &\multicolumn{1}{c}{\bf VL} & \multicolumn{1}{c}{\bf BT} & \multicolumn{1}{c}{\bf MV} & \multicolumn{1}{c}{\bf MD} & \multicolumn{1}{c}{\bf FS} & \multicolumn{1}{c}{\bf EF} & \multicolumn{1}{c}{\bf IT} \\
        \midrule
        Next-Token AR~\citep{DBLP:journals/jmlr/UriaCGML16} & \gcheck & \rcross & \rcross & \rcross & \gcheck & \rcross & \rcross \\
        Permutation-Based AR~\citep{DBLP:conf/icml/UriaML14} & \rcross & \gcheck & \gcheck & \gcheck & \rcross & \gcheck & \rcross \\
        NAR~\citep{DBLP:conf/iclr/Gu0XLS18} & \rcross & \gcheck & \gcheck & \gcheck & \gcheck & \gcheck & \gcheck \\
        Diffusion~\citep{DBLP:conf/nips/HoJA20} & \rcross & \gcheck & \gcheck & \gcheck & \gcheck & \rcross & \gcheck \\
        Consistency Model~\citep{DBLP:conf/icml/SongD0S23} & \rcross & \gcheck & \gcheck & \gcheck & \gcheck & \gcheck & \gcheck \\
        \midrule
        \method (Ours) & \gcheck & \gcheck & \gcheck & \gcheck & \gcheck & \gcheck & \gcheck \\
        \bottomrule
    \end{tabular}
\end{center}
\end{table}

\section{Further Related Work}
\label{sec:related-work}
\paragraph{Order-Agnostic Language Modeling.}
Order-agnostic architectures have been explored to overcome the limitations of sequential generation in autoregressive models. \citet{DBLP:conf/icml/UriaML14} propose permutation-based autoregressive models to learn different data orderings for density estimation. In language modeling, \citet{DBLP:conf/nips/YangDYCSL19} further explore the idea of order-agnostic autoregressive modeling as a generalized pretraining method. \citet{DBLP:conf/icml/WelleckBDC19} explore the possibility of non-monotonic text generation in a tree-structure manner and achieve competitive performance with the conventional left-to-right sequential generation. To avoid the high latency in autoregressive decoding, \citet{DBLP:conf/iclr/Gu0XLS18} introduce non-autoregressive machine translation by breaking the sequentially causal dependency across time into conditionally independent per-step distributions with latent variables as intermediate steps. \citet{DBLP:conf/emnlp/LeeMC18} adopt iterative refinement to interpret the latent variable model, inspired by the design of denoising autoencoders~\citep{DBLP:journals/jmlr/AlainB14}. Follow-up works on non-autoregressive machine translation show promising performance of the iterative refinement process of mask-predict~\citep{DBLP:conf/emnlp/GhazvininejadLL19, DBLP:conf/icml/KasaiCGG20}. Our work explores the potential of unifying the strengths of order-agnostic modeling and denoising to advance sequential modeling in LLMs, demonstrating an efficient way to conduct iterative refinement internally.

\paragraph{LLM Self-Refinement.}
Self-refinement in LLMs focuses on various feedback mechanisms to improve the model performance dynamically. Existing works utilize the feedback mainly in two directions. The first one relates to prompting-based frameworks such as instance-level refinement~\citep{DBLP:conf/nips/MadaanTGHGW0DPY23}, step-level guided search~\citep{DBLP:conf/nips/YaoYZS00N23,DBLP:conf/nips/XieKZZKHX23}, and principle-driven reasoning~\citep{DBLP:conf/iclr/ZhengMCCCLZ24}. Another line of work adapts the feedback as training signals to further enhance the performance of LLMs, including rationale-augmented refinement~\citep{DBLP:conf/nips/ZelikmanWMG22}, hindsight-driven alignment~\citep{DBLP:conf/icml/ZhangLWAG23,zhang2024contrasolverselfalignmentlanguagemodels}, and search-enhanced preference learning~\citep{DBLP:journals/corr/abs-2405-00451}. Unlike existing works relying on the AR foundation in conventional LLMs, we leverage the order-agnostic modeling ability of \method to conduct the iterative refinement internally while foregoing the computation overhead in AR-LLMs to maintain efficiency.

\paragraph{Parallel Decoding.}
Parallel decoding methods aim to accelerate LLM inference by generating multiple tokens simultaneously rather than sequentially. Non-autoregressive models \citep{DBLP:conf/iclr/Gu0XLS18} and blockwise decoding approaches \citep{DBLP:conf/nips/SternSU18, DBLP:journals/corr/abs-2311-13581, DBLP:conf/icml/CaiLGPLCD24} have enabled faster generation but often struggle with output inconsistencies. Speculative decoding techniques \citep{DBLP:conf/icml/LeviathanKM23, DBLP:journals/corr/abs-2302-01318, DBLP:conf/asplos/MiaoOZCWZWZYSSC24} adopts a faster draft model to speedup inference while struggling with the deficiency in scalability. Look-ahead~\citep{DBLP:conf/acl/SantilliSPMMMR23} and Jacobi~\citep{DBLP:conf/icml/FuBS024} decoding, on the other hand, directly utilize the AR LLMs to enhance performance iteratively. Consistency LLMs~\citep{DBLP:conf/icml/KouHHDZ24} further reduces this iteration time drawing inspiration from consistency models~\citep{DBLP:conf/icml/SongD0S23, DBLP:conf/iclr/SongD24}. In this work, we realize parallel decoding leveraging the multi-token generation ability of \method. Instead of decoding toward the forward direction only, we support backward refinement simultaneously to enhance the generation quality further.

\section{Candidate Tree Construction in Order-Agnostic Decoding}
\label{app:decode}
Our specific design of tree construction aims to explore promising combinations of multi-position predictions with a fixed budget for the number of total nodes in the tree. Unlike selecting promising nodes based on the accuracies of the top predictions of different heads in \citet{DBLP:conf/icml/CaiLGPLCD24}, we forego the need of a validation set for accuracy calculation by leveraging the model confidence of each prediction with a dedicated scaling factor. Let $p_{t}^{(i)}$ denote the model-predicted probability of the $i$-th top candidate for the $t$-th token. For a candidate sequence composed by the top $[i_{t_s}, i_{t_s+1}, \cdots, i_{t_e}]$ predictions of tokens at different positions, we estimate its accuracy as:
\begin{equation}
    \prod_{j=t_s}^{t_e}\left(p_{j}^{(i_j)}/\gamma_j\right)
    \label{eq:tree}
\end{equation}
where $\gamma_i$ is a scaling factor to up weight the predictions based on nonconsecutive forward dependencies. As shown in Figure~\ref{fig:accu-loss}, this process benefits from the fact that \method obtains higher accuracies on non-first predictions on such dependencies. Empirically, we set these factors to be $1.1$, $1.2$, $1.3$ for the second, the third, and the fourth tokens to predict, respectively.

Following Eq.~\ref{eq:tree}, we construct the tree in a greedy manner, adding the node with the highest confidence to the tree one by one. This process considers the token-wise confidence as the expected contribution of each prediction to the tree. We repeat the node-adding process until the total number of nodes reaches the desired number to accommodate the maximum sequence length the model can deal with.

\section{Hyperparameter Setting}
\label{app:para}
\paragraph{Training.}
For order-agnostic training, we train for $3$ epochs at each stage with a batch size of $128$ on all tasks. We fix the context window size in training as $4$ and $8$ for forward and backward dependencies. At different training stages, we recommend employing different learning rates. We set the learning rate as $5\mathrm{e}{-6}$ and $1\mathrm{e}{-4}$ in reasoning and code generation, respectively, for the last-layer tuning stage. We increase the learning rates at the second stage to be $1\mathrm{e}{-6}$ and $2\mathrm{e}{-5}$ for corresponding tasks following the general SFT settings.

We corrupt the training data with granularity $4$ and ratio $0.25$ across all tasks for backward reconstruction. As discussed in Section~\ref{sec:ablate}, we ablate the granularity and ratios on mathematical reasoning data to study their respective effects on enhancing model's refinement abilities. Note that as the training context window sizes are fixed as $4$ and $8$ for forward and backward dependencies, the optimal  corruption hyperparameters may vary as we scale the context window sizes. Due to the computation constraint, we leave it to future work to explore the combinations of different granularities and corruption strategies. 

\paragraph{Decoding.}
For order-agnostic decoding, we suggest adopting different context window sizes and block sizes to balance the quality and inference speed in different tasks. We report the experiment results (Section~\ref{exp:rst}) under the same context window and block sizes across the three tasks, where the forward and backward context window and block sizes are $4$, $8$, and $64$, respectively. For verification, we set $\epsilon = 0.2$ and $0.5$ for reasoning and code generation. We implemented our order-agnostic decoding and corresponding next-token baseline without KV-Cache~\citep{DBLP:conf/mlsys/PopeDCDBHXAD23}. During decoding, we set the batch size $1$ and conduct inference on a single GPU.

\paragraph{Computation.}
For reasoning tasks with maximum sequence length $512$, all training experiments were done on single-node eight $40$GB A100s. For code generation task with maximum sequence length $2048$, we conduct training and inference on single-node four $80$GB H100s.

\section{Qualitative Analysis}
\label{app:ext}

\begin{figure}[t]
\begin{center}
\includegraphics[width=\textwidth]{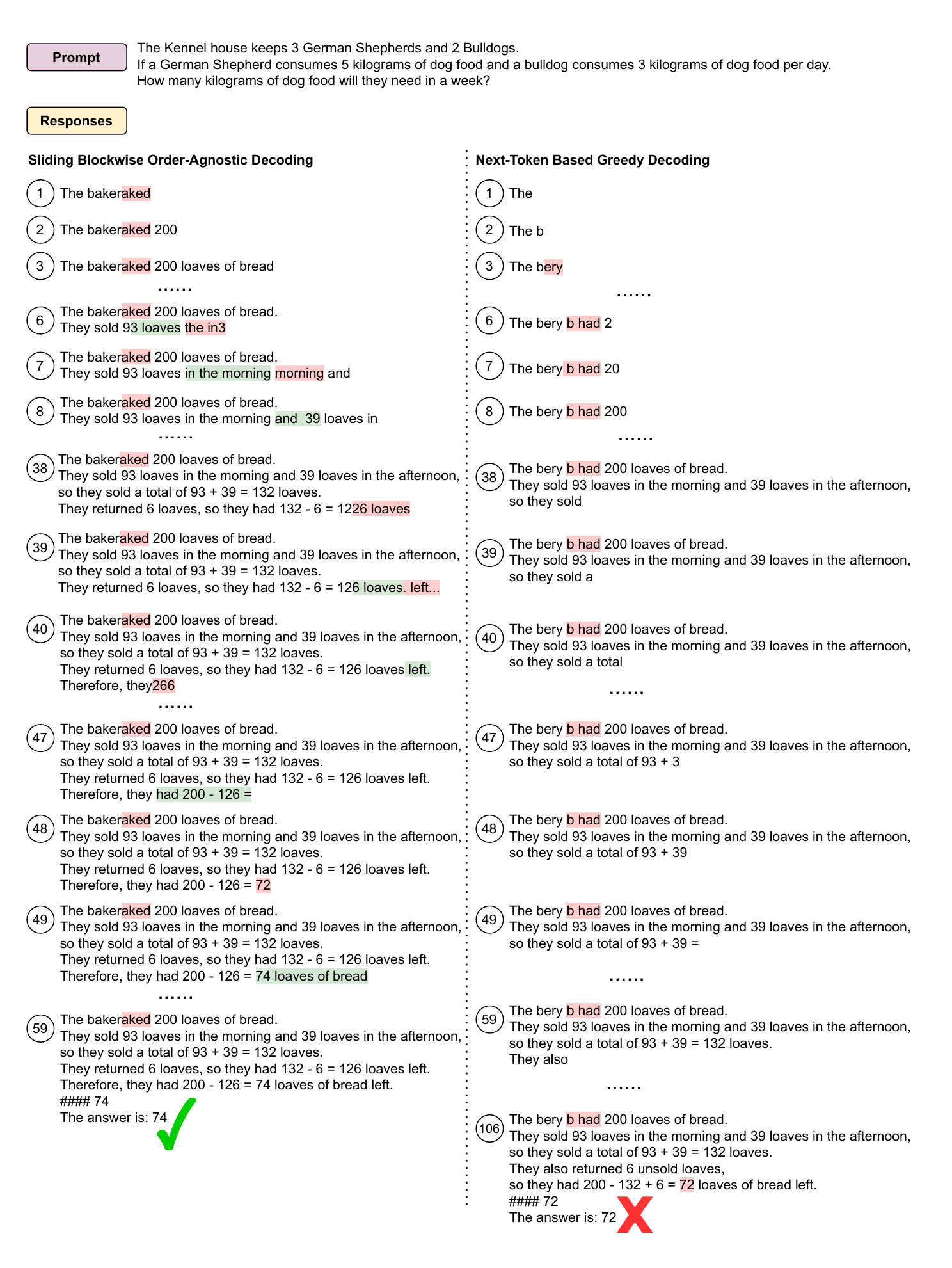}
\end{center}
\caption{Qualitative result comparison on GSM8K.}
\label{fig:case-study-gsm8k}
\end{figure}

\begin{figure}[t]
\begin{center}
\includegraphics[width=\textwidth]{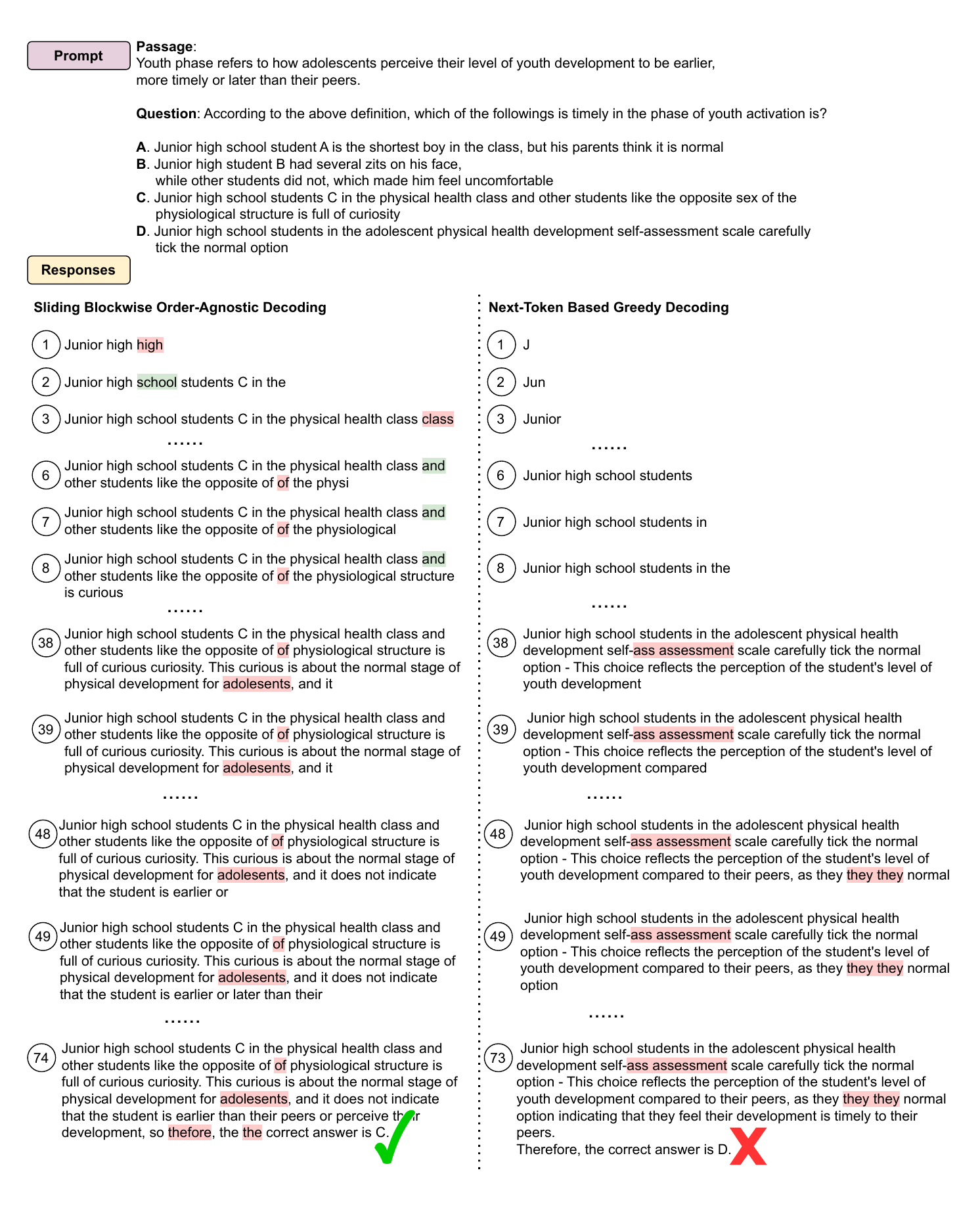}
\end{center}
\caption{Qualitative result comparison on LogiQA.}
\label{fig:case-study-logiqa}
\end{figure}

Our qualitative analysis on GSM8K and LogiQA showcases how \method corrects previously generated mistakes through the iterative internal process. In Figure~\ref{fig:case-study-gsm8k}, \method obtained a wrong calculated result $72$ at the $48$-th step. However, the backward refinement mechanism enables it to backtrack and refine the result to the correct number, $74$, as shown at the $49$-th step. In contrast, the next-token baseline cannot correct the erroneous $72$, leading to the wrong final result. On the other hand, we observe the incoherence in \method's generation where \method can fail in correcting the mistakes when it happens to skip some positions during generation. For example, at the $1$-st step, \method outputs ``bakeraked'' instead of ``baker baked''. This error incurs a chain reaction where the subsequent outputs all omit the correct token `` b'' right after ``baker'', indicating the need for further enhancement on the generation fluency of order-agnostic methods.

On LogiQA, interestingly, we observe a higher frequency of the inconsistencies in \method's generation. As discussed in Section~\ref{exp:rst}, we attribute this scenario to the relatively low proportion of LogiQA-related training data in LogiCoT, where there are only $5$K samples out of the $313$K data points. As shown in Figure~\ref{fig:case-study-logiqa}, while the \method produces several grammatical errors in a generation, it still achieves the correct result. This indicates the advanced ability of \method to sematically escape from paths that may lead to dead ends through iterative refinement.

\end{document}